\documentclass[letterpaper]{article}
\usepackage{natbib,alifeconf}
\usepackage{mdwlist,url}
\usepackage{mathptmx}
\usepackage{moresize}
\usepackage{stmaryrd}
\usepackage{bbm}
\usepackage{amssymb}
\usepackage{graphicx}
\usepackage{verbatim}
\usepackage{subfig}
\usepackage{xcolor,lipsum}

\usepackage{accsupp} 

\makeatletter
\newcommand{\shorteq}{%
  \settowidth{\@tempdima}{-}
  \resizebox{\@tempdima}{\height}{=}%
}

\usepackage{tikz}

\usepackage{calc} 

\usepackage{tikz}

\usepackage[small,compact]{titlesec}

\makeatletter
\newcommand\npar{\@startsection{subsection}{2}{\z@}{-2\p@ \@plus -4\p@ \@minus -4\p@}{-0.5em \@plus -0.22em \@minus -0.1em}{\normalfontnormalsize\bfseries}}
\makeatother

\usepackage{calc} 

\newcommand*\phantomas[3][c]{%
\ifmmode 
\makebox[\widthof{$#2$}][#1]{$#3$}%
\else 
\makebox[\widthof{#2}][#1]{#3}%
\fi 
}

\newcommand{\kleisli}{ \:>\!=\!\!>}

\title{Towards Complex Artificial Life}

\author{Lance R. Williams$^1$\\
\mbox{}\\
$^1$Department of Computer Science, University of New Mexico, Albuquerque, NM 87131\\ williams@cs.unm.edu}

\begin{document}
\maketitle
\begin{abstract}
An object-oriented combinator chemistry was used
to construct an artificial organism with a system architecture
possessing characteristics necessary for organisms 
to evolve into more complex forms.
This architecture supports {\it modularity} by providing a mechanism
for the construction of executable modules called {\it methods} 
that can be duplicated and specialized to increase complexity.
At the same time, 
its support for {\it concurrency} provides the flexibility
in execution order
necessary for redundancy, degeneracy and
parallelism to mitigate increased replication costs.
The organism is a moving, self-replicating, spatially distributed
assembly of elemental combinators
called a {\it roving pile}.
The pile hosts an asynchronous message passing computation 
implemented by parallel subprocesses encoded by genes 
distributed through out the pile like the plasmids
of a bacterial cell.

\end{abstract}

\section{Introduction}

Since its beginning, the field of artificial life has been concerned with the twin
problems of the origin of life on Earth and its evolution into forms of increasing complexity.
Because these problems are among the most important in science, 
the idea that experiments with artificial chemistries, organisms,
and ecologies hosted on computers might substitute for direct observation of events
from the lost history of the early Earth remains extremely seductive.
Still, progress has been slower than many might have expected, 
and artificial life's (arguably) most compelling demonstrations are
already several decades old. 
It follows that a new approach is needed.
In this paper we describe an artificial organism constructed using an object-oriented
combinator chemistry.
While more complex than any previously described, 
it demonstrably possesses a system architecture compatible with
its evolution into still more complex forms.

Phylogenetic reconstructions indicate that all life on Earth descends from 
a last universal common ancestor (LUCA) that existed as early as 3.8 billion years ago \citep{glansdorff}.
This organism was probably a chemical autotroph living near a geothermal vent.
Notwithstanding its likely inability to synthesize amino acids,
it was already quite complex, containing an estimated 355 genes.
Significantly, like all of its descendants, it possessed the molecular machinery
needed to transcribe DNA into RNA, and translate RNA into proteins.
Fossil stromatolites show that by 3.7 billion years ago,
the tree of life rooted at LUCA had branched many times,
yielding a diversity of more complex organisms occupying a 
range of niches in complex ecologies \citep{stromatolite}.


Although the mystery of its origin is paramount among the open questions in our field, 
the question of how an organism of LUCA's non-negligible
complexity evolved into a diversity of still more
complex forms may be more immediately amenable to investigation
using the artificial life approach.
In software engineering terms, did LUCA possess a system architecture that
facilitated its further evolution?
If so, what were the essential characteristics of this architecture?
Could an artificial organism with an architecture possessing these same characteristics be designed?
Would an artificial organism so designed placed in an artificial world where it
was forced to compete with other organisms of the same kind 
for resources evolve into a diverse ecology of still more complex organisms, 
given enough time?
We believe that the answers to the first, third and fourth questions are all `yes'
and these beliefs motivate the present work.
As for the characteristics of LUCA's system architecture that allowed it to evolve 
into more complex forms, 
two of the most likely are discussed in the section that follows.

\section{Accumulation of Complexity}



It has been proposed that a sustained increase in complexity of the most complex
entities of an evolving population is a hallmark of open-ended evolution \citep{york2}.
Although this proposition is compelling, it begs the question of 
how complexity is defined.
In this section, we assume a specific definition for complexity and
describe two classes of mechanisms that together explain
its accumulation in ancient lineages---the first are the source of its increases;
the second mitigate its cost.

The {\it Kolmogorov complexity} of a string is defined as the length of the shortest 
program that prints it.
Unfortunately, Kolmogorov complexity's
usefullness as a measure of the complexity of artificial organisms is limited
because random strings require longer programs
than non-random strings.
A measure that discounts randomness is required.
The {\it logical depth} of a string is the 
{\it time} required to print it given its shortest representation \citep{bennett}.
Because random strings are incompressible,
they are their own shortest representations,
and have low logical depth.

Now consider a string that is a compressed representation of
a decompression program.
When the program is applied to the string, it prints {\it itself.}
It follows that the program plus string system is a {\it quine} with logical depth 
equal to its replication time.
If complexity is equated with logical depth,
then complex organisms require more time than simpler organisms to replicate,
which means that (absent parallelism) complex organisms 
are disadvantaged in zero sum
competitions for resources.


\begin{figure}[t]
\begin{center}
\includegraphics[scale = 0.3, trim = 0 150 0 0,bb = 70 160 700 440]{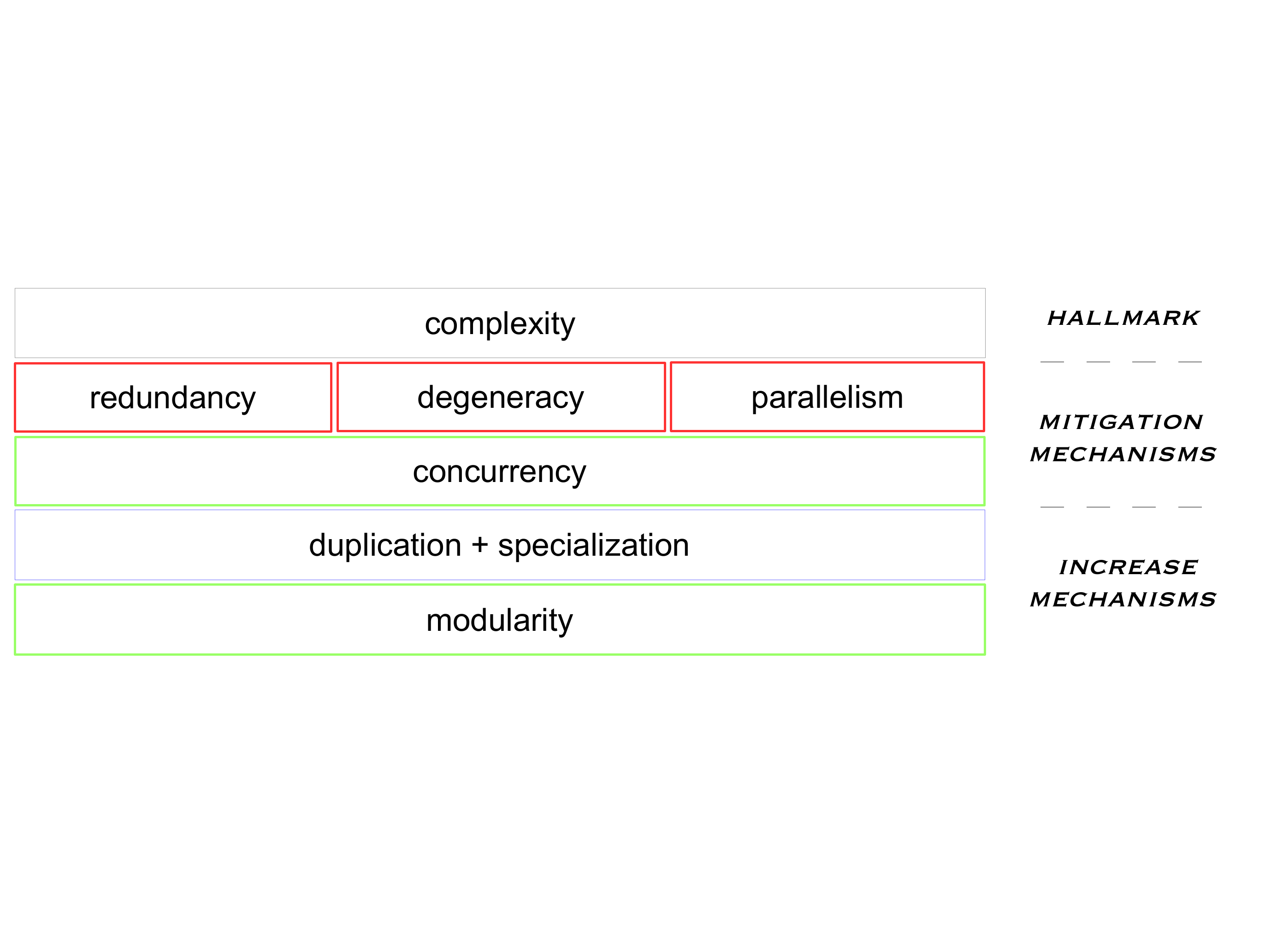}
\caption{
{\it Modularity} facilitates increases in complexity by allowing 
{\it duplication} and {\it specialization} of {\it modules.}
{\it Processes} are executable modules that {\it concurrency} allows to be
executed in different orders.
In a concurrent system, duplication of processes can increase {\it redundancy},
while duplication followed by specialization can increase {\it degeneracy.}
These mitigate the cost of increased
complexity by increasing {\it robustness.}
{\it Parallelism,}
the simultaneous execution of processes on multiple processors, 
mitigates the cost of increased complexity by decreasing the 
time an artificial organism needs to replicate.}
\label{bricks}
\end{center}
\end{figure}

This is a bold claim.
However, it's worth noting that in the natural world,
complex organisms are not intrinsically 
better at staying alive either.
Indeed, 
the theory of {\it constructive neutral evolution} posits
that only the variance in complexity of organisms has increased over time;
its modal value has not \citep{carroll}.
There are innumerably more simple organisms than complex organisms
(no matter how you count)
and organisms as complex as ourselves merely occupy the tail 
of a very broad distribution.\footnote{``That which does not kill us makes us stranger.'' --- Trevor Goodchild, {\it Aeon Flux}.}

Increases in complexity in individual lineages are introduced by evolutionary ``ratchets,''
devices which increase complexity in ways that cannot be reversed \citep{ratchet}.
Although there are others, the most important ratchets 
are {\it duplication} and {\it specialization}.
By means of these devices, complexity accumulates in lineages over time
irrespective of whether or not it confers an adaptive advantage
(see Figure \ref{bricks}).
Sometimes its does; 
just as often it doesn't.

According to this theory,
complex organisms exist primarily due to the fact that life on Earth is ancient.
Generally speaking, they do not survive by virtue of their complexity;
they survive despite it.
For this reason, we believe that an artificial organism capable of open-ended 
evolution must possess 
a system architecture in which both complexity increasing ratchets and factors 
mitigating the costs of complexity increases can be formulated.
The essential characteristics of the system architecture are
{\it modularity} and {\it concurrency}.

Modularity exists at many levels in the biochemical apparatus of the cell.
Protein structural domains, individual proteins, protein
complexes and protein interaction networks have all been described as
``modules''  \citep{modules}. Significantly, there are examples of
increased biological complexity originating from the duplication and 
specialization of modules at each of these levels. 

\begin{figure}[t]
\begin{center}
\includegraphics[scale = 0.3, trim = 0 0 0 0,bb = 100 10 700 400]{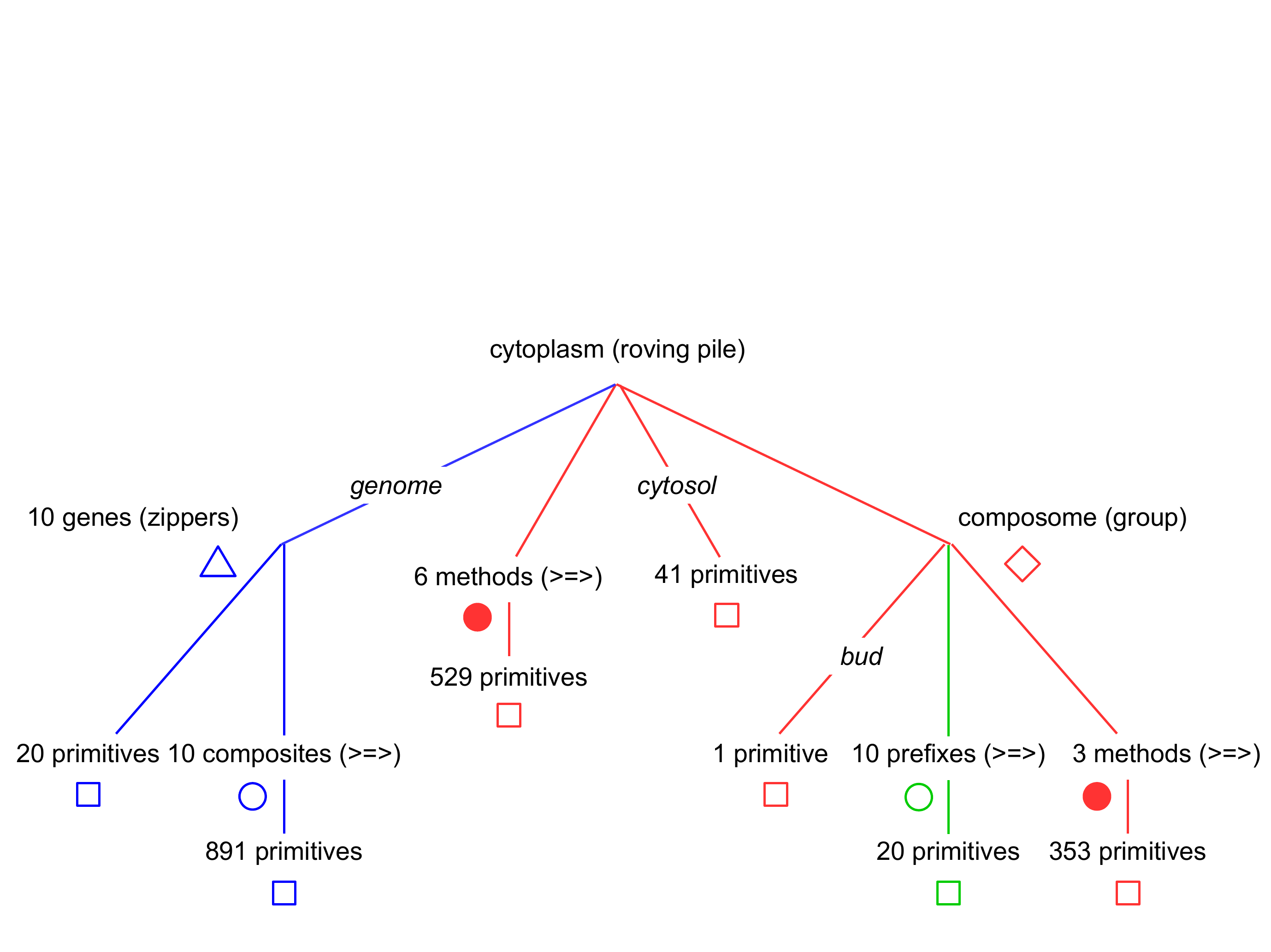}
\caption{The {\it artificial protocell} is a moving, self-replicating, 
spatially distributed assembly of 1855 primitive combinators
called a {\it roving pile}.
Its genome consists of 10 genes represented by {\it zippers} that are distributed 
through out the pile like the {\it plasmids} of bacterial cells.
Methods in the {\it cytoplasm} are executed in parallel and in parallel with 
those in the {\it composome}, which execute concurrently.}
\label{hierarchy}
\end{center}
\end{figure}

If modularity provides the modules that are duplicated and
specialized to increase complexity, then concurrency allows them
to be composed in ways that mitigate the costs of those increases.
Executable modules are {\it processes} and concurrency is the 
property of a system that allows processes to be
executed in different orders without affecting the result. 
More precisely, concurrency allows processes to be executed
in {\it partial orders} defined solely by data dependencies.
This flexibility increases robustness.\footnote{`Robustness'
in the engineering sense, not in the sense 
it is used in evolutionary biology, where it is generally understood
to mean stability of the genotype-to-phenotype mapping.}
While the connection between modularity and evolvability has 
often been emphasized, the importance of concurrency to an
evolvable system architecture has not been previously noted.
This is probably because concurrent execution is the default for biochemical systems.
However, this is not true of computational systems.
Indeed, to our knowledge, there is no artificial organism
apart from our own (see Figure \ref{hierarchy})
that replicates using operations that can be performed in different orders.


A system is {\it redundant} if it contains multiple instances of the same 
component and if working instances can substitute for broken 
instances in the event of failure.
Duplication creates multiple process instances and concurrency
allows one instance to execute instead of another,
yielding redundancy.

A system is {\it degenerate} if it can solve the same problem in different ways \citep{degeneracy}.
Concurrency supports degeneracy because it allows a process derived by
duplication and specialization of an antecedent process to execute 
instead of the antecedent.
Redundancy and degeneracy increase robustness because they allow
organisms to survive component failure and respond in a variety of 
ways to complex environments.

{\it Parallelism} is the simultaneous execution of processes on multiple processors.
Absent a global clock, parallelism is impossible without concurrency;
absent parallelism, complex organisms are at a disadvantage relative
to simpler organisms in the competition for resources,
since they require more time to replicate.



\section{Autocatalytic Set}

{\it Combinators} that return monadic values are the building blocks of
programs in functional programming.
They differ from other notional program building blocks,
{\it e.g.,} {\it bytecodes,} in that monadic
combinators do not require additional address operands to implement
computations which would require statement-level control
in imperative programming, {\it e.g.,} iteration.
Like polypeptides in biochemistry,
programs exhibiting complex behavior can be constructed
from combinators simply by sequencing them.

{\it Object-oriented combinator chemistry (OOCC)} is an artificial chemistry
with composition devices borrowed from object-oriented and functional
programming languages \citep{pop}.
{\it Actors} are embedded in space and subject to diffusion;
since they are neither created nor destroyed, their mass is conserved.
Actors can associate with one another by means of {\it groups} and {\it bonds.}
This allows working sets to be constructed
and the actors in these working sets to be addressed in different ways.
Actors use programs called {\it methods},
constructed from combinators, to asynchronously
update their own states and the states of other actors in their
neighborhoods.
The fact that programs and combinators are themselves reified as
actors makes it possible to build programs that build programs from
combinators of a few primitive types using asynchronous spatial
processes that resemble chemistry as much as computation.


A composite combinator can be represented as a binary tree with
primitive combinators as leaves and interior vertices signifying 
Kleisli composition $(\kleisli)$.
In OOCC, the {\it compose} primitive combinator joins two trees
with  $(\kleisli)$ while the {\it decompose} 
primitive combinator splits a non-leaf tree into 
its two subtrees.
Composite combinators can be promoted to executable
methods using the {\it unquote} primitive combinator.

A {\it zipper} is an implementation of a data structure that allows it
to be traversed and updated without mutation \citep{zipper}.
All zippers consist of three parts. 
The {\it front} represents the portion of the data
structure that has already been traversed, the {\it back} represents
the portion yet to be traversed, and the {\it focus} is a data item
between the front and the back that can be examined or replaced.

A composite combinator's simplest assembly sequence builds it by
adding one primitive combinator at a time via Kleisli composition, 
{\it i.e.}, it is a {\it right fold} with $(\kleisli)$.
This produces a lopsided tree that can be implemented as a {\it list zipper}.
Both the back and the front of the zipper are composite combinators 
with the primitives comprising the front composed in reverse order. 
The zipper's focus is a single primitive combinator.

In a reified implementation in OOCC, a {\it next} bond joins the back and
front while a {\it hand} bond joins the back to the focus.  The zipper
is traversed by pushing the focus onto the front (using {\it compose}), and
popping a primitive combinator from the back (using {\it decompose}).  
This primitive combinator becomes the new focus.

A reversed copy of a composite combinator can be constructed by
traversing its zipper representation.  This is accomplished by
replacing the front with a {\it pair of fronts.}  These are connected
to the back with {\it prev} and {\it next} bonds. At each step of the
traversal, the focus is pushed onto the first front and a primitive
combinator from the neighborhood with type matching the focus is
pushed onto the second front.  This process is repeated until the back
consists of a single primitive combinator, at which point the pair of
fronts represents a reversed original and a reversed copy.  These can
(in turn) be reversed (producing a non-reversed original and
non-reversed copy) by a second traversal of the zipper in the opposite
direction.  This requires creation of a {\it pair of backs}. 
The second back (initially a primitive
combinator from the neighborhood with matching type) is joined to the
first back by making it a member of the first back's {\it group}.\footnote{A group is used instead of a {\it hand} bond so that 
the form of the input to method acsE is distinguishable
from the form of the input to method acsC.}

Note that all of this is accomplished using a very small number of
operations that push (and pop) primitive combinators 
and make (and break) bonds.
Significantly, by using zippers, we eliminate the need for pointers to characters
within arbitrarily long string representations of programs,
{\it e.g.,} as in \cite{hickinbotham}.

\begin{figure}[t]
\begin{center}
\includegraphics[scale = 0.35, trim = 60 120 0 0,bb = 100 130 700 500]{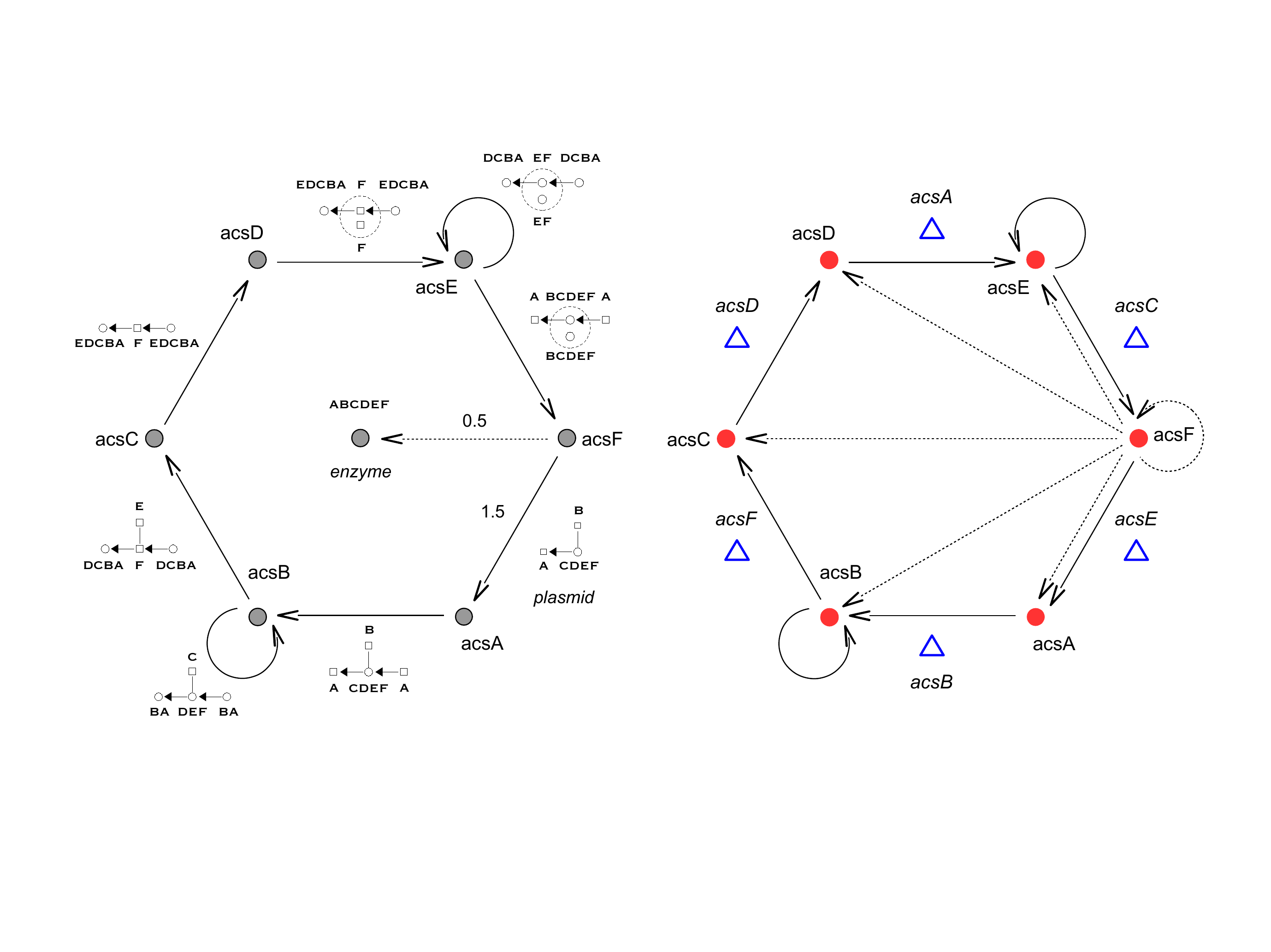}
\caption{Six stage process used to synthesize methods and copy zippers
showing changes to zipper conformation and function of each method
in the parallel pipeline (left).
The addition of six zippers representing the methods implementing the process
itself yields an autocatalytic set (right).}
\label{copying-autocatalytic}
\end{center}
\end{figure}

The copying process is implemented by six methods;
see Figure \ref{copying-autocatalytic} (left).
Initially, the front and focus of
a zipper representing a composite combinator to be copied are both
primitive combinators (the first two forming the composite).
The following operations are performed sequentially:

\begin{itemize}

\item AcsA creates the second front by finding a primitive combinator
matching the front in the neighborhood
and creating the {\it next} bond.

\item AcsB traverses the zipper in the forward direction,
extending the pair of fronts representing 
the reversed original and reversed copy.
At each step, the copy is extended using a primitive combinator
of matching type found in the neighborhood.

\item AcsC adds the final pair of primitive combinators to the pair of
fronts leaving the zipper without a focus and with a back consisting
of a single primitive combinator.

\item AcsD creates the second back by finding a primitive combinator
in the neighborhood matching the first back and joining it to the first back's group.

\item AcsE traverses the zipper in the backward direction, reversing
the pair of fronts by popping primitive combinators off both of them
and pushing these primitive combinators onto the pair of backs.

\item AcsF finishes the reversing process and uses the first back to
construct a zipper in the initial state.  AcsF then (in effect) flips a coin.
If the result is {\it heads,} acsF unquotes the second back, promoting it to a method.
If the result is {\it tails}, acsF uses the second back to a construct a copy of the 
original zipper in the initial state.

\end{itemize}

If the six copying methods acsA--acsF are placed in the world with a
zipper representing a seventh method, then half
of the time, the zipper representing the seventh method will be
copied.  The other half of the time, an instance of the seventh method
will be synthesized.  Once the world contains multiple copies of the
zipper representing the seventh method, the six copying methods will
begin to execute in parallel, forming a production pipeline for inert
(zipper) and active (method) instances of the seventh method.

At this point, an interesting possibility suggests itself.
If the six copying methods acsA--acsF are placed in the world with six
zippers representing the copying methods {\it themselves},
then the twelve entities will form an {\it autocatalytic set} \citep{farmer}.
Over time, the six methods will use the six
zippers to construct additional copies of both methods and zippers.
The methods and zippers are the spatially distributed components of 
a modular, concurrent, parallel, self-replicating system; 
see Figure \ref{copying-autocatalytic} (right).
However, despite these noteworthy attributes,
the autocatalytic set is not a {\it bona fide} artificial organism because it does not segregate
its  components from the components of other systems,
and absent this {\it compartmentalization},
Darwinian evolution is impossible.

\section{Membranes}

After elemental building blocks,
reaction catalysts,
and molecules for storing energy and information,
compartments are probably the next most important ingredient in the recipe for life.  
Given their amazing utility, it is remarkable that, in our universe, 
we basically get them for free.  
This is due to the existence of lipid compounds that, 
when placed in water,
spontaneously assemble into {\it liposomes}, 
vessels defined by bilayer membranes.
Yet membranes are not uncomplicated.
Consider the problem of how to make one grow.
To insert a molecule into a lipid bilayer, a set of forces must be applied
on the lipid molecules adjacent to the point of insertion to create a gap and these
forces must propagate through the bilayer.
They must be combined with the attractive forces the lipids exert on each other
and the forces exerted on the membrane by the cytoplasm.
This {\it mass spring system} requires a physics far more complex 
than the rudimentary one underpinning OOCC,
which has no analog of force.

However, there is a still harder problem associated with growth.
In order for a cell to grow, two different actions must be coordinated.
First, the volume must increase.
This can be done by adding something to the cytoplasm.
Yet if pressure is to remain constant, the membrane 
must also increase in area.
Complicating matters,
the cytoplasm's volume and the membrane's area 
must increase at different rates.
Assuming a spherical cell, an increase in the
volume by $\Delta V$ requires a corresponding increase
in surface area by
\[
\Delta A = \pi (r^3 + \Delta V)^\frac{2}{3} - \pi r^2
\]
\noindent which depends on the cell's radius, $r$.
Given the dependence on $r$, it follows that there is no single local operation that
can maintain constant pressure by pairing imports to both cytoplasm
and membrane. 

Fortunately, membranes are not the only way to achieve the 
compartmentalization necessary for the creation of life.
In fact, in the physical universe, 
a thing as simple as a water droplet in oil can function as a
compartment.
\footnote{\cite{sokolova} have demonstrated transcription
and translation in {\it E. coli} lysate
contained in water-in-oil droplets.}
In a computational universe, a compartment is simply
a data structure for representing a compact, spatially embedded set.
Using a Jordan curve to represent membership in such a set by partitioning space into two
disjoint regions, one ({\it inside}) containing the set's elements,
the other ({\it outside}) containing everything else,
is merely one
possibility.

\section{Roving Piles}

{\it North}, {\it east}, {\it south} and {\it west} are new relations in OOCC
on multisets of actors, or {\it groups.}
We will call the edges of group relations, {\it
links}, to distinguish them from the edges of actor relations, which we call {\it bonds.}
As with actors and bonds, groups can possess at
most one link of each type.
{\it East} and {\it west} are inverse relations, 
{\it i.e.} $E(x,y) = W(y,x)$; the same is true of {\it north} and {\it south}.
Because they correspond to the four cardinal compass directions, links of
these four types are called {\it cardinal links}.
Cardinal links are used to connect {\it base groups}.
{\it Up} and {\it down} are a second inverse relation on groups that can be used to represent
a stack of additional groups above any base group.
A base group is a group without a {\it down} link;
a base group without an {\it up} link is said to be {\it uncovered}.  
A {\it roving pile} is a connected component of base groups embedded in 
the 2D lattice together with the groups contained in stacks above them.
The set of base groups form the pile's {\it footprint}
and base groups with one or more empty 
cardinal links form its {\it boundary.}

In OOCC, methods in the same stack execute concurrently but not in parallel;
they compete for a shared processor resource in zero sum fashion.
However, methods in different stacks in the same pile execute in parallel.
So that piles can move and grow, and so that actors 
within piles can freely mix,
groups in piles are subject to the following three operations:
\begin{enumerate}
\item {\it Diffusion.} A non-base group can be moved to an adjacent stack.

\item {\it Retreat.} An uncovered base group on the boundary can be moved
to an adjacent stack if its removal from the footprint will not split
the footprint into separate connected components.

\item {\it Advance.} A covered base group on the boundary can be replaced
in the footprint by the group above it and used to extend the
footprint in the direction of an empty cardinal link.
\end{enumerate}

Ideally, these operations would be implemented as described above 
and performed at random.  
Unfortunately, the retreat and advance operations,
as described, cannot be implemented using only
local rules.

\begin{figure}[t]
\begin{center}
\includegraphics[scale = 0.3,bb = 75 0 700 200]{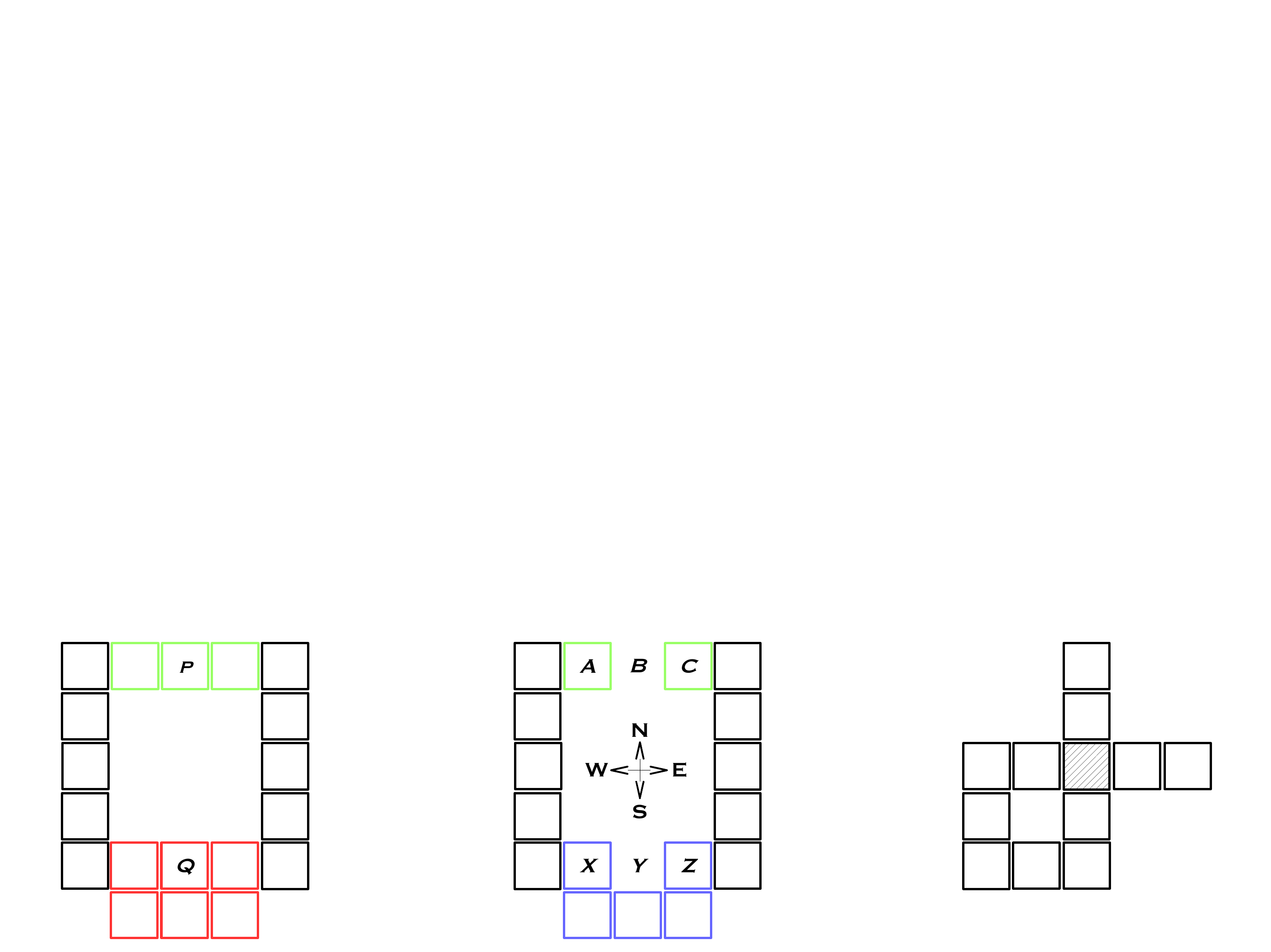}
\caption{Even though it would not split the pile's footprint,
an uncovered base group at P cannot join the stack to its east 
because this cannot be determined by local analysis alone (left).
In contrast, an uncovered base group at Q can do so because it would 
not split the subset of the footprint within its Moore neighborhood (red).
Although a covered base group at A can advance the footprint east
(and C is in the footprint) no link to C will be created (middle).
In contrast, because there is a path between X and Z in the subset of
the footprint contained in the Moore neighborhood of Y (blue),
a covered base group at X can advance the footprint east
and create a link to the base group at Z. 
Because the evolution of roving pile shape is governed solely by 
local rules, pile footprints can overlap (right).
However, actors in 
overlapping neighborhoods cannot interact.}
\label{fig:pile-problems}
\end{center}
\end{figure}

Determining whether or not the removal of a group from the footprint
will split the footprint into separate connected components is
inconsistent with an implementation on an ACA substrate since it is a
function of non-local properties of the cardinal link relation.
For example, the footprint might consist of base groups forming a square
with sides one group wide and $n$ groups long; 
see Figure \ref{fig:pile-problems} (left).
Although it is clear that any single group can be removed without splitting 
the footprint, this can only be determined by traversing a path of length $4n-1$ links.
For this reason, the implementation of the {\it retreat} operation in OOCC
is based on a stronger (sufficient but not necessary) property.
More specifically, an uncovered base group can be removed 
if and only if it will not split the subset of the footprint contained
in its Moore neighborhood into separate components.
This stronger property can be enforced using only local rules.

\begin{figure}[t]
\begin{center}
\includegraphics[scale = 0.3,bb = 100 0 700 250]{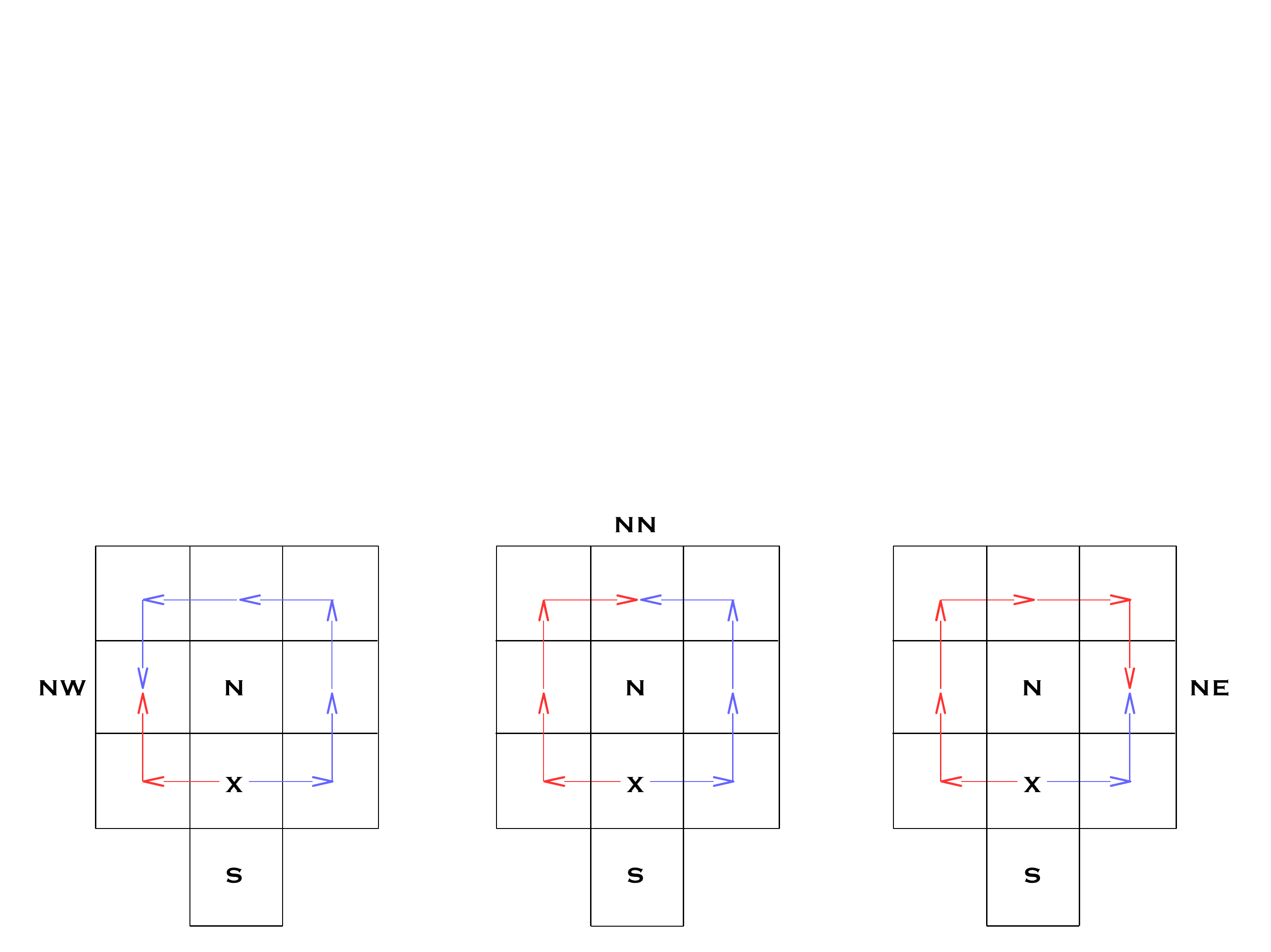}
\caption{
A covered base group at X with an empty {\it north} link can be replaced  in the 
footprint by the group above it in the stack.
Now unlinked, this {\it advance group} can be used to extend the footprint northward to N.
This requires mapping the footprint in the neighborhood of N using search.
Any base groups discovered at NW (left), NN (middle) and NE (right) 
become the advance group's {\it west}, {\it north} and {\it east} links. 
The group that replaced it in the footprint at X
becomes the advance group's {\it south} link.
Corresponding advance operations are performed in the 
other three cardinal directions.
The four together depend only on the topology of the footprint
inside a $5 \times 5$ neighborhood centered on X.}
\label{armpit}
\end{center}
\end{figure}

Implementation of the {\it advance} operation presents a similar problem.
To understand this, consider a roving pile with a square footprint like
the one described above, but with a single group removed;
see Figure \ref{fig:pile-problems} (middle).
In principle, an advance operation could fill the gap, completing the square.
However, this would require a process able to determine whether 
or not base groups adjacent to the advance site are
part of the same pile as itself.  
Again, this can only be done by traversing a path of length $4n-1$ links.  
The solution is to perform an exhaustive enumeration 
within the neighborhood surrounding the advance site; see Figure \ref{armpit}.
This is done to avoid (as much as possible using local rules only), 
the situation where spatially adjacent regions of the
footprint are not connected.

Observations of a working implementation show that roving piles 
remain flat (low average stack height) and connected.
Smaller piles (those containing less than fifty actors) constantly
evolve in shape while rapidly moving around the lattice on random walks.
Holes created by expelling actors in uncovered base groups
are quickly filled.
Larger piles extend and retract pseudopod-like extensions 
but remain largely immobile in aggregate.

Four primitive combinators were added to OOCC to serve as an interface to the roving pile data structure:

\begin{itemize}
\item {\it Safe} fails if the actor it is applied to cannot be removed from the pile without splitting the pile's footprint.
It is used as a {\it guard} for actions that change actors' positions in the pile
or expel actors from the pile. 

\item {\it Expel} removes an actor from the pile.  The actor becomes invisible to actors inside
the pile and visible to actors outside the pile.  This action fails if the actor cannot be expelled
without splitting the pile's footprint.

\item {\it Request} creates a {\it proxy group} representing a request for the pile to import an actor 
of the same type as the actor it is applied to.
It fails if it is applied to an actor which is not a primitive combinator.

\item {\it Seed} creates a new pile containing a single group.

\end{itemize}

\section{Artificial Protocell}

Recent work has described liquid droplets containing enzymes catalyzing growth that 
spontaneously fission into two equal sized droplets upon reaching 
a critical size \citep{zwicker}.
The authors (and others) suggest that droplets like these could form 
the basis of an artificial protocell in {\it vitro}.
The possibility of designing a roving pile with analogous behavior 
that could form the basis of an artificial protocell in {\it silico}
leads us to ask whether an autocatalytic set comprised of method and 
zipper instances of acsA--acsF could be hosted in a roving pile.  
A {\it viable} protocell hosted in a roving pile would contain both 
the autocatalytic set 
and the primitive combinators needed to synthesize it.  
These primitives would be consumed during the process of copying 
methods and zippers, but be replenished by pairing {\it compose} 
actions that consume primitives with {\it request} actions
that replace them while also yielding geometric growth.
This growth would culminate in binary fission.
Assuming that the components of the mother protocell are divided
among its two daughters at random,
then the probability that both daughters will be viable becomes 
closer and closer to one as the mother's size increases.


The approach sketched above seems like a simple and
elegant pathway to an artificial organism possessing modularity, concurrency
and parallelism. 
Unfortunately, there are several practical difficulties.
First, the phenomenon of droplet fission is based on the fact 
that instability increases as droplet size increases.
Because an analogous mechanism devised for roving piles would 
require the computation of the non-local property of pile size,
there can be no {\it simple} mechanism for pile fission.
However, even if a mechanism could be devised,
the pile size of the mother protocell required to reasonably guarantee
the viability of both daughters would still be quite large 
(in the tens of thousands).
For both of these reasons, a different solution was sought.

Absent splitting a mother into two equal-sized daughters,
a daughter must be constructed, 
method-by-method and zipper-by-zipper,
in a process more like {\it budding} than fission.
An efficient construction process would export,
to the daughter, 
one method and zipper instance of each gene,
and the primitive combinators necessary to synthesize both.
To keep track of what has already been exported,
and to recognize when the daughter has
received the full complement of components, 
the mother protocell needs to maintain a checklist of some kind.
We call the group of actors comprising and managing this checklist, 
the {\it composome},
since it serves as the protocell's repository of {\it compositional information.}

The simplest composome would consist of the methods implementing the
export and budding processes, and a set of composites (one per gene)
to represent the checklist.
The copying process in the cytoplasm would translate zippers into composites, 
and each of these would be exported to the daughter as a composite,
method or zipper;
composites exported as composites
would be used to construct the
daughter's composome. 
Composites in the mother's composome would be marked with self-bonds
during the export process to indicate which composites, 
methods and zippers have been exported and
which have not.
After the full complement has been exported,
the bond between mother and daughter (now viable)
would be severed.

Although the approach sketched above works,
it has shortcomings.
First, it is clearly inefficient to use composites
to represent methods and zippers since each has the same length as 
the method and zipper it represents.
Second, requiring two {\it identical} copies of each gene
(a zipper in the cytoplasm and a composite in the composome)
would undermine
evolvability, since a point mutation in either copy would render 
the protocell non-viable.
Recognition of these shortcomings lead to a better approach,
described 
below.

\begin{figure}[t]
\begin{center}
\includegraphics[scale = 0.33, bb = 100 20 700 550]{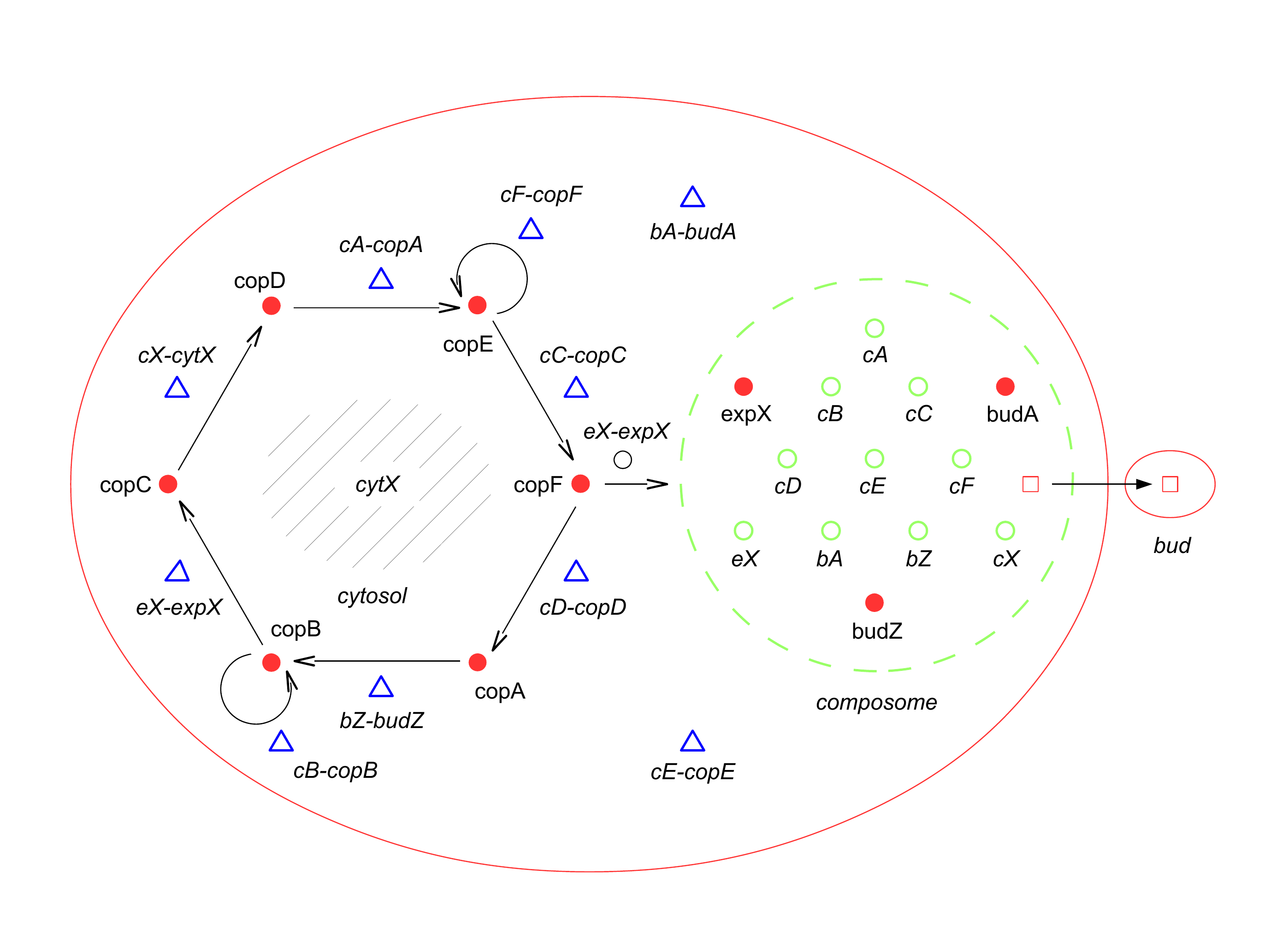}
\caption{Schematic diagram of artificial protocell showing copying of zippers in cytoplasm
and export of zippers and methods by composome through directed bond
to daughter protocell.
Six methods implement the copying process in the cytoplasm,
three methods implement the budding and export process in the composome,
and one method smashes itself to form the cytosol.
The methods in the cytoplasm execute in parallel and in parallel 
with those in the composome (which execute concurrently).
The composome contains ten {\it prefixes}
(length two composites with unique types)
that are marked with self-bonds to signify the zippers and methods
that have already been exported.}
\label{protocell}
\end{center}
\end{figure}

If the composites constructed in the cytoplasm possessed short,
unique, non-executable {\it prefixes,} 
and these prefixes could be used to form the checklist in the composome,
then the protocell would be far more efficient.
Since there is only one copy of each gene (a zipper in the cytoplasm),
evolvability is not undermined; see Figure \ref{protocell}.
This design, for an artificial organism with an architecture 
featuring modularity, concurrency and parallelism, 
has been implemented and tested in OOCC.
It efficiently and reliably replicates across multiple generations 
and possesses only 10 genes:
\begin{itemize}

\item CopA--copE perform operations that are identical to acsA--acsE
except for three small differences.
First, copA--copE are all prefaced by a {\it quit} combinator that is executed 
when the method is exported to the daughter composome.
This causes the method to quit the composome and join the daughter cytoplasm.
Second, copA--copE are modified so that the fronts, focii, and backs of all 
zippers are contained inside single groups.
This avoids the tangling that results when the separate parts of a 
spatially extended zipper joined by bonds occupy different branches of a pile.
Third, all actions that consume primitives in the pile are balanced by
requests to replace them.

\item CopF does the final compose operation needed to complete a
composite representation of a gene for export,
then restores the zipper to the conformation 
expected by the copA method.

\item CytX contains a short executable sequence, $me \kleisli quit \kleisli smash \kleisli none$,
followed by a much longer
non-executable sequence containing one of each of the primitives
necessary for replication (in no particular order).
The short executable sequence causes cytX to quit the daughter composome 
and {\it smash} itself so that the primitive combinators comprising cytX itself 
form the {\it cytosol} of the daughter.


\item ExpX exports composite representations of genes 
as methods and zippers and marks {\it prefixes}
in the mother composome with self-bonds to keep track of progress.
The first two combinators of the composite are removed
and composed to form its prefix.
If the prefix with matching type in the mother composome has no directed self-bond,
then the composite is unquoted and added to the daughter composome
together with its prefix.\footnote{Because unquoted suffixes are methods,
they will execute in the daughter composome when placed there.
Cytoplasm-based methods, {\it e.g.,} copB, are prefaced by 
a pair of combinators, $me \kleisli quit$,
that causes them to quit the daughter composome;
composome-based methods, {\it e.g.,} expX, lack this device.
Like the cytX method used to create the cytosol,
this is a simple use of programmed self-assembly by the daughter.}
If the prefix with matching type has a directed self-bond but no undirected self-bond, 
then the composite and its prefix are used to construct the zipper
representation of the gene and this is added directly to the 
daughter cytoplasm.
Finally, if the prefix with matching type has both directed and undirected
self-bonds, the composite and its prefix are superfluous,
so they are expelled.\footnote{Although OOCC doesn't have
an {\it if-then-else}, equivalent functionality can be achieved
in a single method when all actions are reverseable.
For example, if $z^\prime$ is the action that reverses $z$,
then the sequence, $z \kleisli x \kleisli z^\prime \kleisli y$
will execute the action $z$
when $x$ fails and $y$ when $x$ succeeds.}

\item BudA checks to see if any actor in the composome has a bond.  
If none do, then it expels a primitive from the mother pile and 
applies the {\it seed} combinator to it, creating the daughter pile.
It adds a second primitive to the composome and creates 
a directed bond between it and the first primitive.
Finally, requests are made to the pile to replace both primitives.

\item BudZ checks to see if all prefixes in the composome have directed self-bonds.
If they do, it deletes all prefix self-bonds (directed and non-directed) 
and also deletes the bond connecting the mother and 
daughter, which are now both viable protocells.

\end{itemize}

The artificial protocell is sequential at the top level since it exports methods
and zippers one at a time, as they become available, 
but employs pipeline parallelism in their
production.
There are only two steps in the pipeline that require more than $O(1)$ time.
These are implemented by the copB and copE methods,
which require time proportional to the number of primitive combinators 
comprising the method being copied, $O(M)$.
However, the rate limiting step of the replication process is copB, 
which must wait for the arrival in the neighborhood
of primitive combinators imported by the pile.
It follows that the parallel time complexity of the replication process is
\[
{\textstyle
O\left(\frac{M N}{B}\right) = O\left(\frac{M}{B}\right) \; \sum_{k=0}^{N-1} O \left( \frac{N}{N-k}\right)
}
\]
\noindent where $M$ is the average number of primitive combinators per gene, 
$N$ is the number of genes, and $B$ is the number of instances of copB.\footnote{A breakfast cereal company includes
a plastic dinosaur (one of $N$ different types) in each box of cereal.
It is straightforward to show that a grandmother must buy
$O(N) = \sum_{k=0}^{N-1} O \left( \frac{N}{N-k}\right)$ boxes
on average before her grandson has one of each type.}
Significantly, the time required for self-replication decreases as additional genes
encoding the copB method are added (with diminishing return when $B > N$).
It follows that the protocell is a rare example of a self-replicating  system
where increased  complexity, because it yields increased parallelism, 
pays for itself.

\section{Conclusion}

Because it discounts randomness, computational depth is a useful
measure of an artificial organism's complexity. 
Absent parallelism,
organisms of increased computational depth require more time to replicate.  
This means that they are at a disadvantage relative to simpler organisms 
in zero sum competitions for space.  
It follows that
artificial organisms can only evolve into more complex forms if they
divide the problem of self-replication among 
parallel subprocesses.
In the absence of a
global clock, parallelism is impossible without concurrency, which
allows subprocesses to be executed in different orders. 


Artificial organisms can increase in complexity by means of
duplication and specialization of modules representing subprocesses.
In addition to enabling parallelism, concurrency can mitigate the cost of increased complexity
by providing a variety of execution paths, some of which include these
duplicated and specialized modules.
This can yield increased robustness through redundancy and degeneracy.
We believe that modularity
and concurrency were already present in the cellular architecture of
the last universal common ancestor of all life on Earth and that these
characteristics can be credited in part for its subsequent evolution 
into forms of increased complexity.

Apart from a modular and concurrent architecture, an artificial organism
needs a device for separating its genome and replication machinery
from those of other organisms.
We introduced a new data structure, called a roving pile, capable of representing a set of actors
inhabiting an arbitrarily large four-connected component of sites in a 2D lattice.
Roving piles move and grow and actors within roving piles freely mix,
which is essential for message passing and for the assembly of
methods
from combinators.

Lastly, we used an object-oriented combinator chemistry to construct 
an artificial organism with an architecture featuring
modularity, concurrency and parallelism.
This organism replicates by means of an asynchronous
message passing computation implemented inside of a roving pile 
containing 1855 primitive combinators.
Its genome consists of 10 genes represented by zippers that
are distributed through out the pile like the plasmids
of a bacterial cell.

{\small
\bibliographystyle{apalike}
\bibliography{alife18} 
}

\end{document}